# Multi source feedback based performance appraisal system using Fuzzy logic decision support system

GMeenakshi

Asst.Prof,Nalla Malla Reddy Engineering College, Department of Computer Science
Hyderabad,Andhra Pradesh,India

*meena_ganti@yahoo.com*

## Abstract

*In Multi-Source Feedback or 360 Degree Feedback, data on the performance of an individual are collected systematically from a number of stakeholders and are used for improving performance. The 360-Degree Feedback approach provides a consistent management philosophy meeting the criterion outlined previously. The 360-degree feedback appraisal process describes a human resource methodology that is frequently used for both employee appraisal and employee development. Used in employee performance appraisals, the 360-degree feedback methodology is differentiated from traditional, top-down appraisal methods in which the supervisor responsible for the appraisal provides the majority of the data. Instead it seeks to use information gained from other sources to provide a fuller picture of employees' performances. Similarly, when this technique used in employee development it augments employees' perceptions of training needs with those of the people with whom they interact. The 360-degree feedback based appraisal is a comprehensive method where in the feedback about the employee comes from all the sources that come into contact with the employee on his/her job. The respondents for an employee can be her/his peers, managers, subordinates team members, customers, suppliers and vendors. Hence anyone who comes into contact with the employee, the 360 degree appraisal has four components that include self-appraisal, superior's appraisal, subordinate's appraisal student's appraisal and peer's appraisal .The proposed system is an attempt to implement the 360 degree feedback based appraisal system in academics especially engineering colleges.*

## 1. Introduction

In recent years multi-source feedback systems (MSFS) also known as 360 Degree Appraisal became very popular. It became popular as it has been felt for long years that one person's assessment of another individual cannot be free of biases. In addition, with the focus on customers (both internal and external) and emphasis on softer dimensions of performance (leadership, innovation, team work, initiative, emotional intelligence, entrepreneurship etc.) it

                                                     

has become necessary to get multiple assessments for a more objective assessment. 360 Degree Appraisal is Multi- Rater Appraisal and Feedback System. Almost every Fortune 500 Company is using this in some form or the other. In this system, the candidate is assessed periodically (once in a year and sometimes even half yearly) by a number of assessors including his boss, immediate subordinates, colleagues, internal customers and external customers. The  assessment is made on a questionnaire specially designed to measure behaviors.

Performance appraisal the latest mantra for career development is followed by many organizations across the world. "Get paid according to what you contribute" this is turning the focus of organization to performance management and individual performance. It helps to rate the employees and evaluate their contribution towards the organizational goals based on their performance. "*Free Form method" generally involves* description of the performance of an employee by his superior. This is an evaluation of the performance of any individual based on the facts and often includes examples and evidences to support the information. This system has the inseparability of the bias of the evaluator as major drawback. To overcome this new form of feedback "360-degree feedback" is formed, it is also known as 'multi-rater feedback'. In this system the feedback is taken from all the sources which come in contact with the employee on his/her job. The various sources include co-workers, managers and supervisors, customers, staff and the individual being evaluated. This provides full assessment of an individual based on feedbacks from multiple sources.

It is an evaluation tool utilizing opinions of many different people who interact with the employee on a routinely manner. It generates more accurate feedback by gathering information from people about individual's performance as seen by the organizational structure & expectations of their boss, self, peers, subordinates & customers. Definition takes different form when this system is applied to engineering education. Here participants are principal, head of the department, teaching staff members, students & laboratory assistants.

In industries, 360 degree performance appraisal system is widely used nowadays. In 1997, only 8% industries were using it, while this percentage has gone up to 52% by 2008. As far as engineering education is concerned (India & Abroad), this percentage is very less (upto 12%). Implementing 360 degree performance appraisal system provides more comprehensive performance ratings, as employees are given an opportunity to map their competencies. Performance rating of teacher can be decided by taking views of principal, head of the department, teaching staff members, students, laboratory assistants & university results. This appraisal system fits well in an educational institute. It will be a big task to measure individual's skills, competencies, motivational drivers, work habits & potential for developing future competencies precisely. It is the best tool for identification of strengths of staff members for career development. It also identifies weaknesses for training & it can be used for salary recommendations. Feedback can be obtained by using a questionnaire which asks participants to rate the individual according to observed competencies/behaviors & data.

## 1.1 The 360 degree performance appraisal system

Typically, performance appraisals have been limited to a feedback process between employees & superiors. With the increased focus on teamwork, employee development & customer service, the emphasis has shifted to employee feedback from the full circle of sources. This multi input approach to performance feedback is called "360 degree assessment" to connote that full circle as shown in fig. 1.

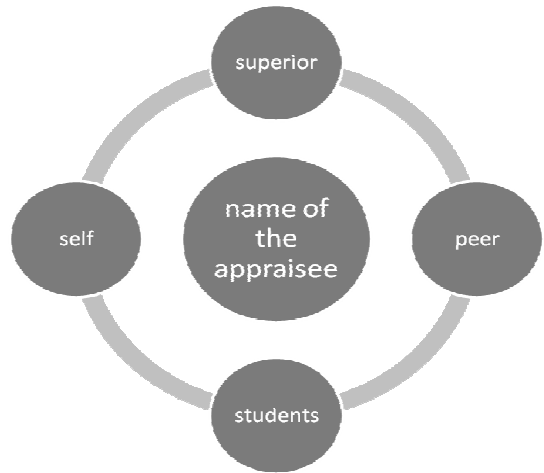

Figure 1 : 360 degree sources of feedback

This system is a holistic approach incorporating views from many angles, multi level & multi source appraisal. Now by changing focus from industry to academia, sources in the circle will change. Different methods are available to assess the performance. Proper questionnaire has to be designed. For a teaching staff member, feedback from principal, students, colleagues, HOD & lab assistant will play an important role. Different methods are available to assess the performance. Subject results should be compared with the university results. While taking feedback from students, rating of students should also be decided. Following aspects are important for teaching staff member: Subject matter Mastery Contribution to curriculum development, Instructional designs & delivery, establishing a positive learning environment completing related administrative requirements, Community partnership includes developing partnerships with individuals, groups, social organizations outside the Institution.

## 2. Related Works

### 2.1. Performance Appraisal System

Performance appraisal systems are quite useful in understanding & assessing the skills, potential and productive output of an employee. There are many methods available today for evaluating employee performance; 360 degree evaluation is an effective way of evaluating the performance of an employee. Like any other method, 360 degree appraisals need to be carried out with care to obtain a fair and an accurate result. Harinder Singh, HR & strategic head, Vigneshwara Developers, tells us about the method, "360 degree appraisal is a comprehensive method wherein the feedback about the employees' performance comes from all the sources that come in contact with the employee on his job. The respondents for an employee can be his/her peers, managers, subordinates, team members, customers, suppliers/vendors; hence, anyone who comes into contact with the employee and can provide valuable insights and feedback regarding the 'on-the-job' performance of the employee. 360 degree appraisal has four integral components that include self-appraisal, superior's appraisal, subordinate's appraisal and peer appraisal." 360 degree appraisals are found to be very effective in assessing an employee's performance.

Vishal Chibber, director HR, Kelly Services India, tells us about the benefits of this method, "360 degree feedback is one of the most widely used employee assessments today as the process is more transparent and beneficial compared to any other appraisal procedure."

In a developing & knowledge-based economy, it is very important for organizations to understand the competencies needed in the workforce for business success, and then develop those qualities & skills on an ongoing basis. The appraisal results are used to identify the better performing employees who should get the majority of available merit pay increases, bonuses, and promotions.

Performance appraisal is a formal management system that provides for the evaluation of the quality of an individual's performance in an organization [4]. As mentioned by Dessler, G [5], performance appraisal has the means to evaluate an employee's current and past performance relative to the employee's performance standards. It is a process which involves creating work standards; evaluate employee's actual performance relative to those work standards; and giving feedback to employee so as to motivate him or her to improve the job performance or to eliminate performance deficiency. In addition to that, Terrence, H. M and Joyce, M. [6] stated that, some potential aims of performance appraisal might include identifying particular behavior or job. Various techniques or methods have been used by human resource management experts to evaluate the performance of an employee. As outlined by Vicky G. [7], some of the appraisal methods include ranking; trait scales; critical incident; narrative; and criteria-based. Terrence, H. M. and Joyce, M. [6] mentioned few other methods including management-by-objectives (MBO), work planning and review, $360^o$ appraisal and peer review. With all the available techniques, it is essential to understand that different organization might use different technique in assessing staff performance. Since all the techniques mentioned above has their own advantages and disadvantages, most organizations might mix and match different techniques for their own performance appraisal system that can fulfill their organizational needs. Performance appraisal system has become one of the most valuable management tool in which organization members use to achieve collective goals. In order to ensure that the results of the performance appraisals are useful and reasonable to the superior when evaluating their subordinates, it is important for the performance appraisal system to consistently produce reliable and valid results for the management of an organization. Fuzzy based method has been applied into several performance appraisal systems. Proposed a methodology utilizing fuzzy set theory and electronic nominal group technology for multi-criteria assessment in the group decision-making of promotion screening. The study suggested that the methodology is a good method for a transparent and fair multi-criteria performance evaluation in military organizations. Researchers have demonstrated that fuzzy set theory could be successfully used to solve multiple criteria problems [8]. This is because, in many circumstances, appraiser tends to use vaguely defined qualitative criteria in evaluating the performance of their subordinates. Therefore, it creates difficulty for appraiser to precisely quantifying the score of each candidate. worked on applying fuzzy set theory on computer-based fuzzy group decision support system (FGDSS). Based on the findings of their work, the application of fuzzy set theory in FGDSS is said to be able to assist decision maker to make better decisions under different circumstances and alternatives a good example of the application of the fuzzy-set theory to decision-making process is multifactorial evaluation model. This study has suggested that information can reasonably obtain and saw ability classification is reasonable and acceptable. The literatures that have been reviewed supported that the fuzzy set theory would be a good concept to be used in the development of the performance appraisal system. This is because fuzzy set theory allows the performance appraisal system to be developed by using some fuzzy variables and relationships. Therefore, the idea of incorporating this model in the performance appraisal system can be a promising approach.

## 2.2. Application of Fuzzy based Multifactorial Evaluation Method

Multifactorial evaluation is a good example of the fuzzy-set theory to decision-making process. Its purpose is to provide a synthetic evaluation of an object relative to an objective in a fuzzy-decision environment that has many factors. Let U={$u_1,u_2$ ,$u_3$,........ $u_n$) be a set of objects for evaluation., let F={$f_1,f_2,f_3$,.......$f_m$} be the set of basic factors in the evaluation process, and let E = {$e_1$, $e_2$,......, $e_n$} be a set of descriptive grades or qualitative classes used in the evaluation. For every object u € U there is a single- factor evaluation matrix R(u) with dimension m x p, which is usually the result of a survey. This matrix may be interpreted and used as a 2-D MF for fuzzy relation F X E.

With the preceding three elements, F,E and R, the evaluation result D(u) for a give n object u € U can be derived using the basic fuzzy processing procedure: the max-min composition of fuzzy relations, where the resulting evaluation is in the form of a fuzzy set D(u) = [ $d_1$, $d_2$, $_3$, $d_4$] :

$$D(u) = W(u).R(u) = [0.4\ 0.4\ 0.2] = \begin{bmatrix} 0.6 & 0.2 & 0.1 & 0.1 \\ 0.1 & 0.5 & 0.3 & 0.1 \\ 0.1 & 0.3 & 0.4 & 0.2 \end{bmatrix}$$

$$=[\ 0.4\quad 0.4\quad 0.3\ 0.2\ ]$$

where example d1 is calculated through the following steps:

$$d1 = (d1 = (w1 \wedge r11) \vee (w2 \wedge r21) \vee (w3 \wedge r31)$$
$$=( 0.4 \wedge 0.6 ) \vee (0.4 \wedge 0.1 ) \vee (0.2 \wedge 0.1)$$
$$= 0.4 \vee 0.1 \vee 0.1$$
$$= 0.4$$

the values for d2 d3 and d4 are found similarly, where ^ an V represent the operations min and max ,respectively. Because the largest components of D(u) are d1 = 0.4 and d2 = 0.4 at the same time, the analyzed piece of faculty receives a rating somewhere between SUPERIOR AND EXCEPTIONAL.

# 3. Proposed Methodology

Before the actual implementation of the system, questionnaires were prepared and distributed to the human resource section to evaluate the usability and effectiveness of the system. In order to demonstrate the application of multifactorial evaluation model in the performance appraisal system multifactorial evaluation model was developed. This performance appraisal system uses al criteria provided by the company by using the following steps.

- Design questionnaire
- Feedback process
- Evaluation & communication
- Formation of developmental plans
- Follow up
- Survey.

## 3.1 PERFORMANCE APPRAISAL MODEL

*T*o reward and develop the human resource of the organization Performance appraisal is used by an organization to ensure that the organization runs smoothly and grow. Staffs are required to fill up every year, Yearly/semester. Work plans prepared to report on the progress of the tasks assigned as agreed early of the year. This is done at the beginning and at the middle of the year. At year end, the Yearly Work Plan is used to evaluate the performance of the staff throughout the whole year. The following diagram describes the performance appraisal model of the company can be implemented in academics also as shown in Figure 2

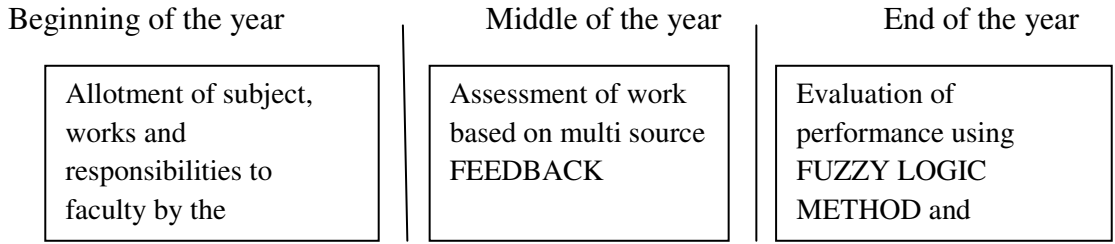

Beginning of the year          Middle of the year          End of the year

| Allotment of subject, works and responsibilities to faculty by the | Assessment of work based on multi source FEEDBACK | Evaluation of performance using FUZZY LOGIC METHOD and |

Fig.2

## 3.2 Staff Evaluation

Four aspects will be taken into consideration when evaluating staff performance and each aspect will index into its sub criteria, as follows: audit and helps in revamping HR processes and systems for improved performance. Performance appraisals are frequently used in organization as a basic for administrative decision such as employee promotion, transfer and allocation of financial rewards; employee development, including identification of training needs and performance feedback.

Aspect 1: Planning & Preparation of course (Effectiveness & Punctuality): designing student centric instructional content/course objectives, selecting instructional goals, assessing student learning.

Aspect 2: The class room Environment (maintaining class room discipline & control of class): Creating an environment of respect and rapport, establishing a culture for managing class room procedures and managing student behavior.

Aspect 3:Instuction delivery:(Communication & Knowledge skills) Communicating very clearly and accurately, using questioning and discussion techniques engaging student in learning providing feedback to students, demonstrating flexibility and responsiveness.

Aspect 4: Professional Responsibilities& contributions (towards college and society): Maintaining accurate records, communication with families contributing to the college and state, growing and developing professionally.

(a) Working output (Aspect 1): This aspect evaluates the quantity, quality and effectiveness of the staff's working output as well as staff's punctuality.

(b) Knowledge and skills (Aspect 2): This aspect evaluates the staff's knowledge and skills in the working field as well as their effectiveness in communication and realization of rules.

(c) Personal quality (Aspect 3): This aspect evaluates the personal quality appreciated by the organization such as discipline, proactive, innovative, cooperativeness and independence.

(d) Informal Event(s) and Contribution(s) (Aspect 4): Staff's contribution to the organization, community, state, country and international.

When evaluating staff's performance, appraiser will use a scale of 1 to 10 to rate each sub criteria for each aspect mentioned above. 1 indicates that the staff was rated poorly in that particular sub criteria and 10 indicates that the staff was rated highly in a particular sub criteria. The verbal grade for the scale is shown in table 1.

VERBAL GRADES AND SCALE FOR EACH ASPECT

| VERBAL GRADES | SCALE |
|---|---|
| EXCEPTIONAL | 9 or 10 |
| SUPERIOR | 7 or 8 |
| FULLY SUCCESSFUL | 5 or 6 |
| MINIMALLY SUCCESFUL | 3 or 4 |
| SATISFACTORY | 1 or 2 |

TABLE I

## 3.3 Performance appraisal Process

The proposed Model , the performance appraisal system is a combination of four multifactorial evaluation models.

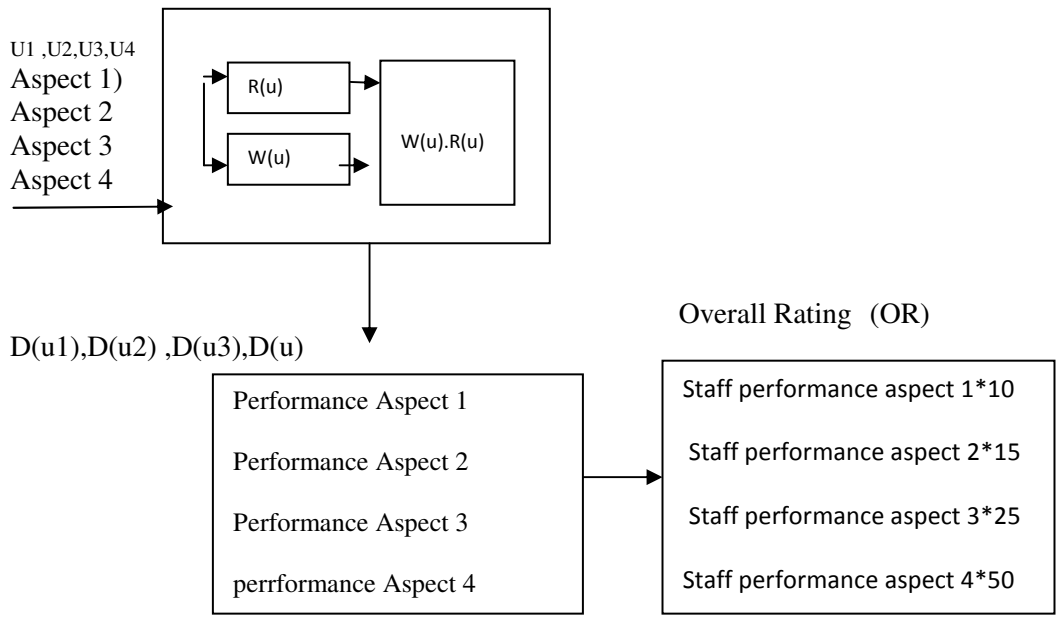

Fig. 3 Proposed Application of Multifactorial Evaluation Model in Performance Appraisal System

Algorithm:

Step: 1.Use linguistic weighting variables.
Step:2.Use linguistic rating variables
Step: 3.Linguistic evaluations converted to matrix to construct fuzzy decision Matrix.
Step: 4.The normalized fuzzy decision matrix is a single factor evaluation vector.
Step: 5. Multiplication of matrices W(u) and R(u) is based on the min-max composition of fuzzy relations.

As shown in Fig. 3, the models represent aspects to be evaluated in the staff performance appraisal. After getting the performance on each aspect from their superior, the staff's overall ratings can be calculated by the following equation shown in Fig. 3 above. In Fig. 3, $U$ is the factors to be evaluated in each aspect whereas $D(u)$ is the result of staff's performance in a particular aspect. The first evaluation model in Fig. 3 uses $U1$, that is, the factors in Aspect 1 as its input.

The sub criteria for this aspect will be used as the basic factors under this aspect which are:
f1 = quantity of working output
f2 = quality of working output
f3= Punctuality and
f4 = Effectiveness of working output

Therefore, F = {$f_1$, $f_2$, $f_3$, $f_4$}.

The verbal grades used for the appraisal are:
$e_1$ = Exceptional
$e_2$ = Superior
$e_3$ = fully successful
$e_4$ = minimally successful

$e_5$ = Satisfactory

Therefore, E = {$e_1$, $e_2$, $e_3$, $e_4$, $e_5$}.

The single-factor evaluation for each aspect for a staff's performance in terms of working output(Aspects), can be derived as follows. As an example, the weights of the f1 are, 10% for Exceptional, 40% for superior, 30% for Fully successful, 10% for minimally successful, and 10% for satisfactory, thus, the single-factor evaluation vector R1(u) is:

R1(u) = {0.1, 0.4, 0.3, 0.1, 0.1}

In the same way, the single-factor evaluation vectors for f2, f3, and f4 which have been gained are as shown as below:-

R2(u) = {0.2, 0.5, 0.2, 0.1, 0.0}

R3(u) = {0.5, 0.3, 0.1, 0.1, 0.0}

R4(u) = {0.2, 0.5, 0.2, 0.1, 0.0}

As a result, by referring to the single-factor evaluations vectors stated above, the following evaluation matrix can be built:-

R(u) =     R1(u) = {0.1 0.4 0.3 0.1 0.1}
           R2(u) = {0.2,0.5 0.2 0.1 0.0}
           R3(u)={ 0.5 0.3 0.1 0.1 0.0}
           R4(u)= {0.2 0.5 0.2 0.1 0.0}

W(u) is the weighting factors. Therefore, it is the appraiser's rating towards a staff for all the sub criteria in a particular aspect. As an example, assume that the appraiser's rating for weight vector corresponding to the four factors in all aspects are:

W1(u) = {0.2, 0.3, 0.3, 0.2}

W2(u) = {0.3, 0.4, 0.3}

W3(u) = {0.2, 0.3, 0.2, 0.3}

W4 (u) = {0.1}

Multiplication of matrices W(u) and R(u) is based on the min-max composition of fuzzy relations, where the resulting evaluation is in the form of a fuzzy set D(u) = [d1, d2, d3, d4] [9]. Since the aspect of Working Output has five verbal grades, that is,

E = {e1, e2, e3, e4, e5}

which would be involved in the performance appraisal system, thus, the resulting evaluation is in the form of a fuzzy set D(u) = [d1, d2, d3, d4,d5] as shown below:-

$$D(u) = W1(u) \cdot R(u) = [0.2\ 0.3\ 0.3\ 0.2] \cdot \begin{bmatrix} 0.1\ 0.4\ 0.3\ 0.1\ 0.1 \\ 0.2\ 0.5\ 0.2\ 0.1\ 0.0 \\ 0.5\ 0.3\ 0.1\ 0.1\ 0.0 \\ 0.2\ 0.5\ 0.2\ 0.1\ 0.0 \end{bmatrix} = [0.3\ 0.3\ 0.2\ 0.1\ 0.1$$

They are calculated through the following steps as shown below whereby ^ represent the operations min and v represent the operation max.

d1 = (w1 ^ r11) v (w2 ^ r21) v (w3 ^ r31) v (w4 ^ r41)
= (0.2 ^ 0.1) v (0.3 ^ 0.2) v (0.3 ^ 0.5) v (0.2 ^ 0.2)
= 0.1 v 0.2 v 0.3 v 0.2
= 0.3

d2 = (w1 ^ r12) v (w2 ^ r22) v (w3 ^ r32) v (w4 ^ r42)
= (0.2 ^ 0.4) v (0.3 ^ 0.5) v (0.3 ^ 0.3) v (0.2 ^ 0.5)
= 0.2 v 0.3 v 0.3 v 0.2
= 0.3

d3 = (w1 ^ r13) v (w2 ^ r23) v (w3 ^ r33) v (w4 ^ r43)
= (0.2 ^ 0.3) v (0.3 ^ 0.2) v (0.3 ^ 0.1) v (0.2 ^ 0.2)
= 0.2 v 0.2 v 0.1 v 0.2
= 0.2

d4 = (w1 ^ r14) v (w2 ^ r24) v (w3 ^ r34) v (w4 ^ r44)
= (0.2 ^ 0.1) v (0.3 ^ 0.1) v (0.3 ^ 0.1) v (0.2 ^ 0.1)
= 0.1 v 0.1 v 0.1 v 0.1
= 0.1

d5 = (w1 ^ r15) v (w2 ^ r25) v (w3 ^ r35) v (w4 ^ r45)
= (0.2 ^ 0.1) v (0.3 ^ 0.0) v (0.3 ^ 0.0) v (0.2 ^ 0.0)
= 0.1 v 0.0 v 0.0 v 0.0
= 0.1

Since the largest components of

$$D(u) \text{ are d1} = 0.3, \text{ and d2} = 0.3$$

at the same time. Referring to the verbal grades,

E = {EXCEPTIONAL, SUPERIOR, FULLY SUCCESSFUL, MINIMALLY SUCCESSFUL, SATISFACTORY}

the analyzed staff's performance in terms of working output obtained a rating somewhere between "EXCEPTIONAL" and "SUPERIOR". However, by applying the principle of the biggest subjection degree as used by Guifeng, G. *et. al.* [12], the staff's performance in terms of working output is Aspect is"EXCEPTIONAL".

# 4. Experimental Results

Table II: Verbal grades and weighting for each aspect of performance

| Aspect | Verbal Grades | Weighting for Each Aspect |
|---|---|---|
| **Aspect 1** | Exceptional | 1.0 |
| | Superior | 0.8 |
| | Fully Successful | 0.6 |
| | Minimally Successful | 0.4 |
| | Satisfactory | 0.2 |
| **Aspect 2 & 3** | Exceptional | 1.0 |
| | Superior | 0.8 |
| | Fully Successful | 0.6 |
| | Minimally Successful | 0.4 |
| | Satisfactory | 0.2 |
| **Aspect 4** | Exceptional | 1.0 |
| | Superior | 0.8 |
| | Fully Successful | 0.6 |
| | Minimally Successful | 0.4 |
| | Satisfactory | 0.2 |

The same method of calculation can be applied to $U2$, $U3$, and $U4$, which are, the aspect of Knowledge and Skills, the aspect of Personal Quality, and the aspect of Informal Event(s) and Contribution(s), respectively. Following this, the verbal grades and weighting for each aspect as shown in Table II is being referred when calculating a staff's overall average ratings:-The staff's overall average rating (AR) is:-

AR = (Aspect 1 * 50) + (Aspect 2 * 25) + (Aspect 3 * 20) +(Aspect 4 * 5)Based on Table II, the analyzed staff's performance in terms of working output obtained a rating of "EXCEPTIONAL". As a result, 1.0 would be the weighting for Aspect 1. Meanwhile, according to what have been computed by using the multifactorial evaluation model, the staff has been rated as "FULLY SUCCESSFUL" or the weighting of 0.6 in terms of Knowledge and Skill. As for the aspect of Personal Quality, the staff's performance is "MINIMALLY SUCCESSFUL" or the weighting of 1.0 would be selected. As for the Informal Event(s) and Contribution(s), the staff has gained an "SATISFACTORY" performance or the weighting of 0.8 for this aspect. Thus, the rating and weighting for each aspect is as summarized below:-TABLE III

Table-III: SUMMARIZED RATING AND WEIGHTAGE OF THE 4 ASPECTS of Performance

| Aspect | Rating | Weighting |
|--------|--------|-----------|
| Aspect  1 | EXCEPTIONAL | 1.0 |
| Aspect 2 | SUPERIOR | 0.6 |
| Aspect 3 | FULLY SUCCESSFUL | 1.0 |
| Aspect 4 | SATISFACTORY | 0.8 |

Table-IV: Marks for different sources of  feedback

| s.r.no | Feed back | Marks |
|--------|-----------|-------|
| 1. | *Feedback from superior* | 50 |
| 2. | Feedback from peer groups | 25 |
| 3. | Feedback from students/results | 20 |
| 4. | Self | 5 |

Here we have following points to remember:

• Every source of feedback  has marks.
• According to select option, final getting marks  of particular  source  is  calculated  using above table grades and scale.
Therefore, the staff's Overall Performance Rating (OPR) is:-

OPR = ((1.0 * 50) + (0.6 * 25) + (1.0 * 20) + (0.8 * 5))= 89

fig 4:

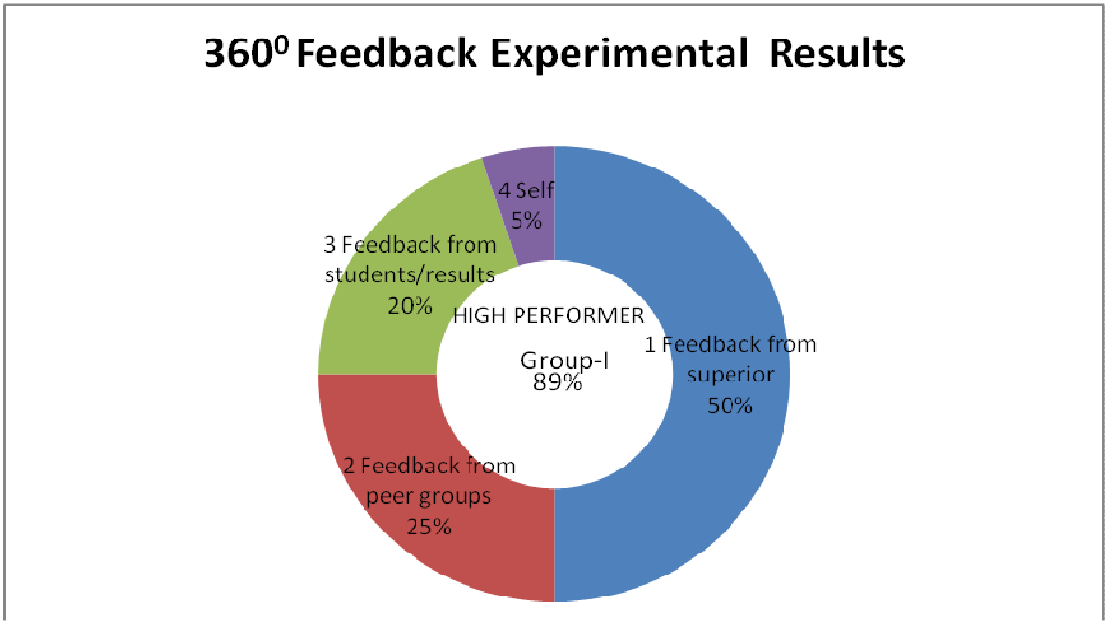

As a result, according to the calculation above and by referring to Table V below, the staff would be categorized in the "High Performer(s)" group, Medium Performer(s) group, Average Performer(s) group and Low Performer(s) group.

Table V BENCHMARK / STANDARD USED IN DETERMINING STAFF'S PERFORMANCE

| Overall Average Ratings | Group | Remarks |
|---|---|---|
| Above 80%<br>(GROUP I) | High performer(s) | • An incentive of RM 1000.<br>• A certificate of appreciation.<br>• Entitled for "Best Service Award". |
| Less than 80%<br>but more than 60<br>(GROUP II) | Medium Performer(s) | • An incentive of RM 500<br>• Advised to improve their performance in the coming year |
| Less than 60%<br>but more than 50%<br>(GROUP III) | Average performer(s) | • Advised to improve their performance in the coming year<br>•should attend training sessions& workshops |
| Less than 40%<br><br>(GROUP IV) | Low Performer(s) | • Disciplinary action<br>*might be* taken towards the staff.<br>• Should constantly report his / her work progress to his / her assessors in a stated period |

Fig 5:    BENCHMARK / STANDARD USED IN DETERMINING STAFF'S PERFORMANCE

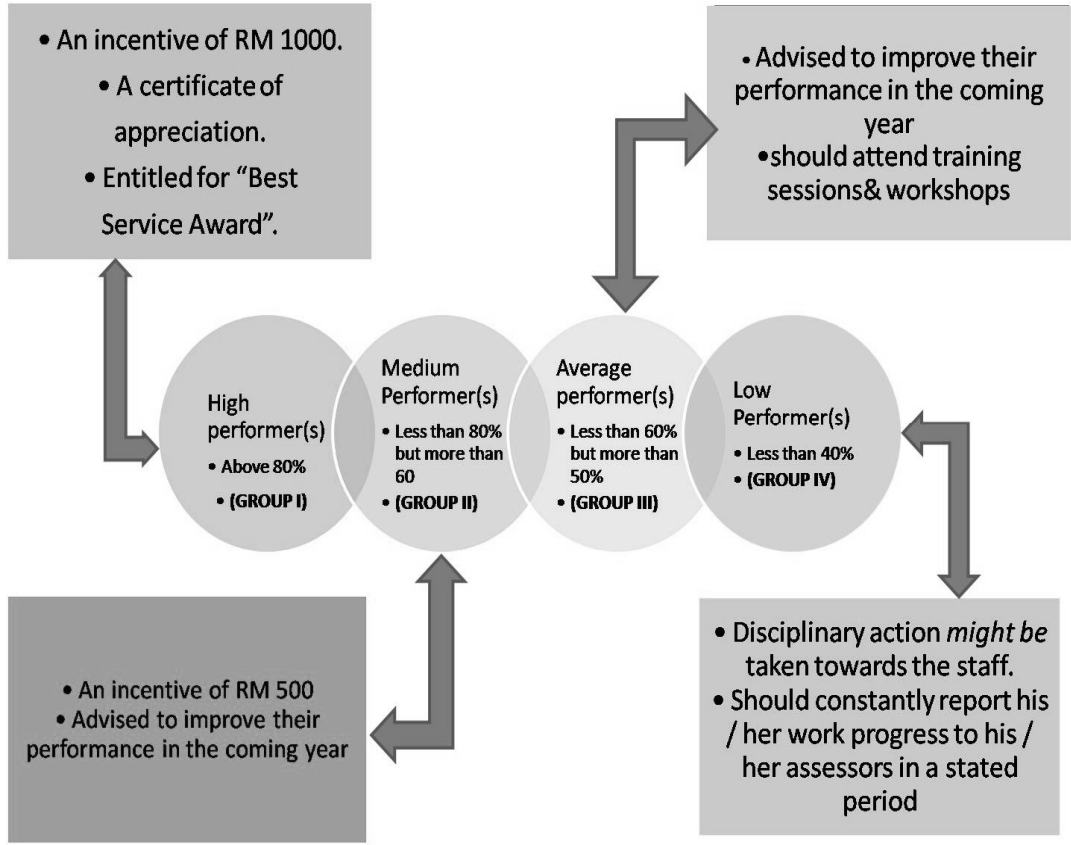

## 5. Conclusion & Future direction

Multifactorial evaluation model is used in assisting high-level management, to appraise their employees. Utilizing the concept of using four multifactorial evaluation model in the performance appraisal system could ease the changes need to be made in this system whenever it is necessary. This model follows a systematic step in determining a staff's performance, and therefore, it creates a system of appraisal which is able to consistently produce reliable and valid results for the appraisal process. In order to allow others to use this system, the aspect to be evaluated and the weightage for each of these aspects need to be define in the system before hand.

Good engineering institutes are those where activities are designed & prompted which result in personal, social, academic & career oriented growth of students & staff. 360 degree performance appraisal is the best tool to achieve this goal. Objective of this system is to identify areas for both organizational & individual improvement. For educational institutes, it is necessary to align personal goal of staff member with organizational expectation. This system can be used as a diagnostic & assessment tool to increase employee participation & to demonstrate a commitment to their workforce. 360 degree feedback determines relationship between strategic plan/vision of institute &performance expectations. It increases focus on customer service & reinforce continuous process improvement programs. Team based culture can be developed for attaining organizational objectives. Staff members become more inclined to consider factors beyond

HOD's expectations when exhibiting behaviors & striving for results. It detects barriers to success.

For performance assessment and adequate support in decision making the proposed model produced significant bases. So the research on the issue can be continued. Important aspect of this issue that could focus on in the future is that this model can be extended to all types of employee assessment in Universities and Engineering colleges. For fast results Excel charts can be used or an interface can be developed using math-lab or vc++.

## 6. Acknowledgement

I am heartily thankful to Dr. Mrs. Divya Nalla , Principal, Nalla Mallareddy Engineering College Whose encouragement and support resulted in the preparation of paper.I am thankful to Dept HOD for allotting us research hours. I am also thankful to Mr. Mani Sarma Assoc. Prof whose encouragement guidance and support from the initial to the final level.

## Author

Author's Profile:G.Meenakshi received her M.C.A degree in 2001 and M.Tech degree in 2010 and presently working in the department of computer science as Asst.prof in Nalla MallaReddy Engineering College Hyderabad Andhra Pradesh.Her research araea is Artificial Intelligence(Fuzzy logic).

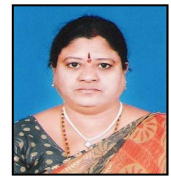